\documentclass[conference]{IEEEtran}
\IEEEoverridecommandlockouts
\usepackage{cite}
\usepackage{amsmath,amssymb,amsfonts}
\usepackage{algorithmic}
\usepackage{graphicx}
\usepackage{textcomp}
\usepackage{xcolor}
\def\BibTeX{{\rm B\kern-.05em{\sc i\kern-.025em b}\kern-.08em
    T\kern-.1667em\lower.7ex\hbox{E}\kern-.125emX}}

\usepackage{url}
\usepackage{graphicx}
\usepackage{epstopdf}
\usepackage{footnote}
\usepackage{balance}
\usepackage{amssymb}
\usepackage{amsmath}
\usepackage{multicol}
\usepackage{multirow}
\usepackage{anyfontsize}
\usepackage{array}
\usepackage{hhline}
\usepackage{float}
\usepackage{enumitem}
\usepackage{tabularx}

\begin{document}

\title{OmniTrack: Real-time detection and tracking of objects, text and logos in video\\
}

\author{\IEEEauthorblockN{Hannes Fassold, Ridouane Ghermi}
\IEEEauthorblockA{\textit{JOANNEUM RESEARCH - DIGITAL} \\
Email: {firstname.lastname}@joanneum.at}
}


\maketitle
\begin{abstract}
The automatic detection and tracking of general objects (like persons, animals or cars), text and logos in a video is crucial for many video understanding tasks, and usually real-time processing as required. We propose OmniTrack, an efficient and robust algorithm which is able to automatically detect and track objects, text as well as brand logos in real-time. It combines a powerful deep learning based object detector (YoloV3) with high-quality optical flow methods. Based on the reference YoloV3 C++ implementation, we did some important performance optimizations which will be described. The major steps in the training procedure for the combined detector for text and logo will be presented. We will describe then the OmniTrack algorithm, consisting of the phases preprocessing, feature calculation, prediction, matching and update. Several performance optimizations have been implemented there as well, like doing the object detection and optical flow calculation asynchronously. Experiments show that the proposed algorithm runs in real-time for standard definition ($720x576$) video on a PC with a Quadro RTX 5000 GPU.
\end{abstract}
\begin{IEEEkeywords}
object detection, tracking, YoloV3, optical flow
\end{IEEEkeywords}

\section{INTRODUCTION}
\label{sec:intro}  

The automatic detection and tracking of general objects (like persons, animals or cars), text and logos in a video provides semantic information which is crucial for many high-level computer vision tasks in various application areas ranging from visual surveillance, autonomous driving to automatic video annotation and brand monitoring. In most of these application areas, it is essential that the detection and tracking is done in real-time (\emph{online tracking}). 

In this work we present OmniTrack, a runtime-efficient and high-quality algorithm which is able to automatically detect and track general objects (which we identify with the 80 classes from the MS COCO dataset \footnote{\url{http://cocodataset.org}}), text as well as brand logos in real-time. It combines a sophisticated deep learning based object detector with a high-quality GPU-accelerated optical flow method, which makes the algorithm very robust even for difficult content. The rest of the paper is organized as follows: In Section \ref{sec:related} some related work is presented. In Section \ref{sec:detector} we describe the performance optimizations we applied to the reference C++ implementation of the YoloV3 object detector and provide details of the training procedure for the combined text and logo detector. In section \ref {sec:flow} we outline the employed optical flow engine. Section \ref{sec:algorithm} describes the overall OmniTrack algorithm, its main phases and performance optimizations we implemented for the algorithm. Section \ref{sec:experiments} describes result regarding the performance (runtime and quality) of the algorithm for content of different type and resolution, and Section \ref{sec:conclusion} concludes the paper.

\section{RELATED WORK}
\label{sec:related}

All real-time capable approaches in the literature seem to focus either on general objects, text or logos, but not on the detection and tracking of \emph{all three categories at once} (as our proposed approach does). We therefore report related work from each of the three categories separately. For general objects, the approach of\enspace\cite{Bewley2016} employs a combination of a deep learning based object detector (Faster R-CNN), Kalman filters for prediction (tracking) and simple association techniques. The work\enspace\cite{Luo2018} proposes a detect or track (DorT) framework. In this framework, object detection/tracking of a video sequence is formulated as a sequential decision problem where a scheduler network makes a detection/tracking decision for every incoming frame, and then these frames are processed with the detector/tracker accordingly. With respect to text, the method in\enspace\cite{Gomez2014} achieves real-time performance by combining a text detection component based on MSER features with a tracking component based on the MSER component tree. Note that the usage of hand-crafted features (like MSER or SIFT) is deprecated nowadays because their performance is inferior to learned features based on neural networks. The work\enspace\cite{Yang2018} describes an online video text detection method which works nearly in real time. It detect texts in each frame using the EAST algorithm, and formulates the online text tracking problem as decision making in Markov Decision Processes (MDPs). With respect to logos, in\enspace\cite{Natarajan2011} a method is reported for large-scale, real-time logo detection and recognition in broadcast video. The algorithm identifies first a small set of possible logo locations in a frame, based on temporal continuity and multi-resolution search, and then successively prunes these locations for each logo template, using a cascade of color and edge based features. The runtime is approximately $100$ milliseconds per frame on a PC with a 3.0 GHz CPU.

\section{YOLOV3 OBJECT DETECTOR}
\label{sec:detector}

The YoloV3 object detector \cite{Redmon2018} is a state of the art deep learning based algorithm, which provides a very good compromise between detection capability and runtime. It processes the input image in a single phase, in contrast to other popular approaches like Faster-RCNN, which work in two phases (generation of region proposals, classification of regions). The algorithm divides the image into a fixed $n \times n$ grid. For each cell of the grid, it predicts a fixed number of bounding boxes together with their confidence scores and class IDs. A custom 53-layer network called Darknet-53 is employed.

For the robust and efficient detection of general objects (persons, animals, cars, ...), text and logos in the image, we propose to employ two different network models. The first one is responsible for the detection of general objects (80 object classes), and employs the standard YoloV3 network model trained on the MS COCO dataset and which is publicly available. The second is responsible for the detection of both text and logos (two object classes). As such a network model is not publicly available, we trained a custom network model for this task. In Section \ref{subsec:training_custom} the dataset generation and training procedure will be described in detail. Employing one network model for both text and logo, instead of two separate models, is significantly more runtime-efficient as it saves one full inference pass. We employed several other important performance optimizations ((adaptive size of the receptive field, multiple inference calls on the same GPU in parallel via worker threads) which we describe in Section \ref{subsec:yolov3_optimization}.

\subsection{Custom network model for text and logo detection}
\label{subsec:training_custom}

\begin{figure}[t]
	\centering
		\includegraphics[width=0.4\textwidth]{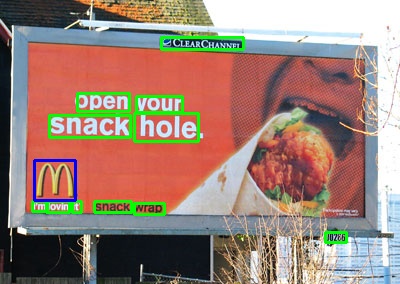}
	\caption{Example of combined logo \& text detection on a test image from the Logos in the Wild dataset~\cite{Tuzko2018}, green corresponds to text detections and blue to logo detections.} 
	\label{fig:yolov3_example_text}
\end{figure}

 For the detection of text and logos, no publicly available models exist which detect both at once. Thus, we decided to train our own model on a custom dataset created by merging the \emph{COCO-Text} dataset \footnote{\url{https://bgshih.github.io/cocotext/}} and the \emph{Logos in the Wild} dataset \cite{Tuzko2018}. The COCO-Text dataset is a large-scale scene text dataset based on the MSCOCO. We used the version 2.0 which contains 63,686 images along with 239,506 annotated text instances.On the other hand, the Logos in the Wild Dataset is a large-scale set of images with logo annotations. The version we used (v2.0) consists of 11,054 images with 32,850 annotated logo bounding boxes of 871 brands.  Our resulting dataset contains 74,740 images along with 272,356 annotations. There are 10 times more text annotations than logo annotations. However, this is not a problem because the YoloV3 model shows resilience to class imbalance. The last layer of the YoloV3 model is a bunch of independent sigmoid neurons, so, even if the two classes can be very similar (e.g. logos can contain text), training for one class will not affect the other class. One important parameter to tune in the YoloV3 model is the size of the anchor boxes. YoloV3 is a one-stage detector which estimates the bounding box coordinates of an object not from scratch but by refining pre-defined bounding boxes called anchor boxes. The choice of those anchor boxes is crucial because it will give strong baselines to the model to learn a specific task. Therefore, we use k-means clustering (with $k = 9$) to extract the 9 most relevant bounding box sizes from our dataset.  We train our model from ImageNet pre-trained weights with the standard training settings of YoloV3 on batch of images resized to $608x608$ during 50k iterations on a Nvidia Quadro P5000. We reach a loss of $0.84$ with $0.44$ mAP. Figure \ref{fig:yolov3_example_text} shows an example of detection on a test image.

\subsection{Performance optimizations}
\label{subsec:yolov3_optimization}

Since the reference C++ implementation from \footnote{\url{https://github.com/pjreddie/darknet}} employs only a fixed-size receptive field of quadratic size, an important modification to the implementation has been done in order to support adaptive receptive fields. The idea is that the receptive field size (which is quadratically in the reference implementation) is adapted dynamically, so that it has  the same aspect ratio as the input image. This makes sure that the receptive field of the neural network is exploited in the most efficient way. For an Full HD ($1920x1080$) image with $16:9$ aspect ratio, this means that the size of the receptive field will be $608x342$ instead of the standard size of $608x608$ pixel (of which roughly $40\%$ of the pixels would be wasted due to the proportional resizing of the input image so that it fits into the receptive field).

Another limitation of the reference C++ implementation is that only \emph{one} inference call is done at a time on a specific GPU. This is due to the usage of a single CUDA stream and CUDNN handle for that GPU (both are global variables), leading to a serialization of the CUDNN library calls in the reference implementation. Obviously, this leads to a underutilization when running the inference on modern GPUs which provide a massive amount of computational resources. In order to get around this limitation with minimal modifications of the reference code, we change these global variables to thread-local variables so that each CPU thread has its own private CUDA stream and CUDNN handle. This modifications makes it on the other hand necessary that all calls for a certain network instance come from the \emph{same} CPU thread so that always the same CUDA stream and CUDNN handle is used by this instance. Therefore, we create for each network instance its own persistent worker thread, which is responsible for making all the calls (instance creation/deletion, inference, ...). These modifications make it possible to do multiple inference calls (in our case two inferences - one for the standard model for MS COCO and one for our combined model for text/logo detection) in parallel on the same GPU. The parallelization of the two inference calls brings brings an runtime improvement of approximately $20\%$ on a Quadro RTX 5000 GPU.
\section{OPTICAL FLOW ENGINE}
\label{sec:flow}

The optical flow problem deals with the calculation of a dense pixel-wise motionfield between two images and has many important applications, as in our case the prediction of the object position in the next frame (tracking) within the OmniTrack algorithm. We employ a GPU implementation of the the TV-L1 algorithm  \cite{Werlberger2009} , a classical variational algorithm. It formulates the problem in a disparity preserving, spatially continuous way with a L1 norm data term and anisotropic TV regularization, employing the Huber Norm  and solving it numerically with a projected gradient descent scheme.
The GPU optical flow implementation is programmed in CUDA  and runs in real-time for Full HD resolution on a decent GPU (like a Geforce GTX 1070). The input images for the optical flow algorithm are preprocessed via a structure-texture decomposition in order to be more robust against brightness variation (e.g. due to flicker) in the video. 

\section{OMNITRACK ALGORITHM}
\label{sec:algorithm}

The design of the OmniTrack algorithm  is inspired by recent real-time capable online multi-object tracking algorithms like \cite{Bewley2016}, which employs a combination of a deep learning based object detector (Faster R-CNN), Kalman filters for prediction and simple association techniques. The OmniTrack algorithm employs more powerful components instead, namely the state of the art YoloV3 object detector \cite{Redmon2018} and a dense optical flow algorithm for prediction of the object movement. In the following, we will outline the main phases of the overall algorithm and describe the performance optimizations we have implemented for it in order to make it real-time capable.

\subsection{Overall algorithm}
\label{subsec:overall}

In the following, the overall workflow of the OmniTrack algorithm is described briefly. Let $I_{t-1}$ and $I_t$ be two consecutive frames for timepoints $t–1$ and $t$, and $S_{t-1}$ be the list of already existing scene objects which have been tracked so far until timepoint $t–1$. 
The workflow for each frame can be roughly divided into four phases: 
\begin{itemize}[leftmargin=0.2in]
	\item In the \emph{preprocessing} phase, some general preparation steps are done for the input image $I_t$ (the current frame of the video). Specifically, the input image is downscaled via a gaussian pyramid, and a proper pyramid level is chosen based on the input image resolution. We employ downscaling for performance reasons, as it would be very computationally expensive to process e.g. 2K or 4K resolution images in their original resolution.
	\item In the \emph{feature calculation} phase, the optical flow between the (downscaled) frames $I_{t-1}$ and $I_t$ is calculated, which gives a dense motion field $M_{t-1}$.  Additionally, the YoloV3 object detector is invoked for $I_t$ with two different network models, yielding a list of detected objects $D_t$. The two network models employed are the standard model for detecting the 80 MS COCO classes and our custom model we trained for the detection of both text and logos.
\item In the \emph{prediction} phase, for all scene objects $S_{t-1}$ their motion is predicted in order to calculate their predicted position $S_t$ for timepoint $t$. The prediction for each object is calculated as the average of the motion vectors in $M_{t-1}$ for the object bounding box.
\item In the \emph{matching} phase, a matching is done of the predicted scene objects $S_t$ against the detected objects $D_t$. The matching is done in a globally optimal way with the Hungarian algorithm~\cite{Kuhn1955} , and the \emph{IOU} (intersection-over-union) metric is employed as the matching score.
\item In the \emph{update} phase, all scene objects in $S_t$ which could not be matched are considered as lost and a flag is set to exclude them from further processing. In contrast, all detected objects in $D_t$ which could not be matched are considered as newly appearing and therefore added to the scene objects list $S_t$. 
\end{itemize}

\subsection{Performance optimizations}
\label{subsec:omnitrack_performance}

In order to utilize both CPU and GPU as much as possible, we parallelized several steps in the algorithm which are independent of each other. A good candidate for parallelization is the motion field calculation and the object detection (both done in the feature calculation phase), as these steps do not have mutual dependencies. Therefore, we run both steps asynchronously in different CPU threads, which brings a runtime improvement of roughly $20\%$ for a video with a resolution of $2560x1280$ pixel.

Additionally, we do also some phases for $I_{t}$ and $I_{t+1}$ (so for two consecutive frames) in parallel. This kind of \emph{prefetching} introduces a lag of one frame, as not only the current frame $I_{t}$ has to be provided, but also the next frame $I_{t+1}$. Prefetching allows use to run all algorithm phases for the frame $I_{t}$ in parallel with the preprocessing phase (pyramid calculation, ...) for the frame $I_{t+1}$. By this we effectively hide the computational expense for the preprocessing phase which brings another speedup of the OmniTrack algorithm of $10-15\%$.

\section{EXPERIMENTS AND TESTS}
\label{sec:experiments}

\begin{figure*}[t]
	\centering
		\includegraphics[width=0.75\textwidth ]{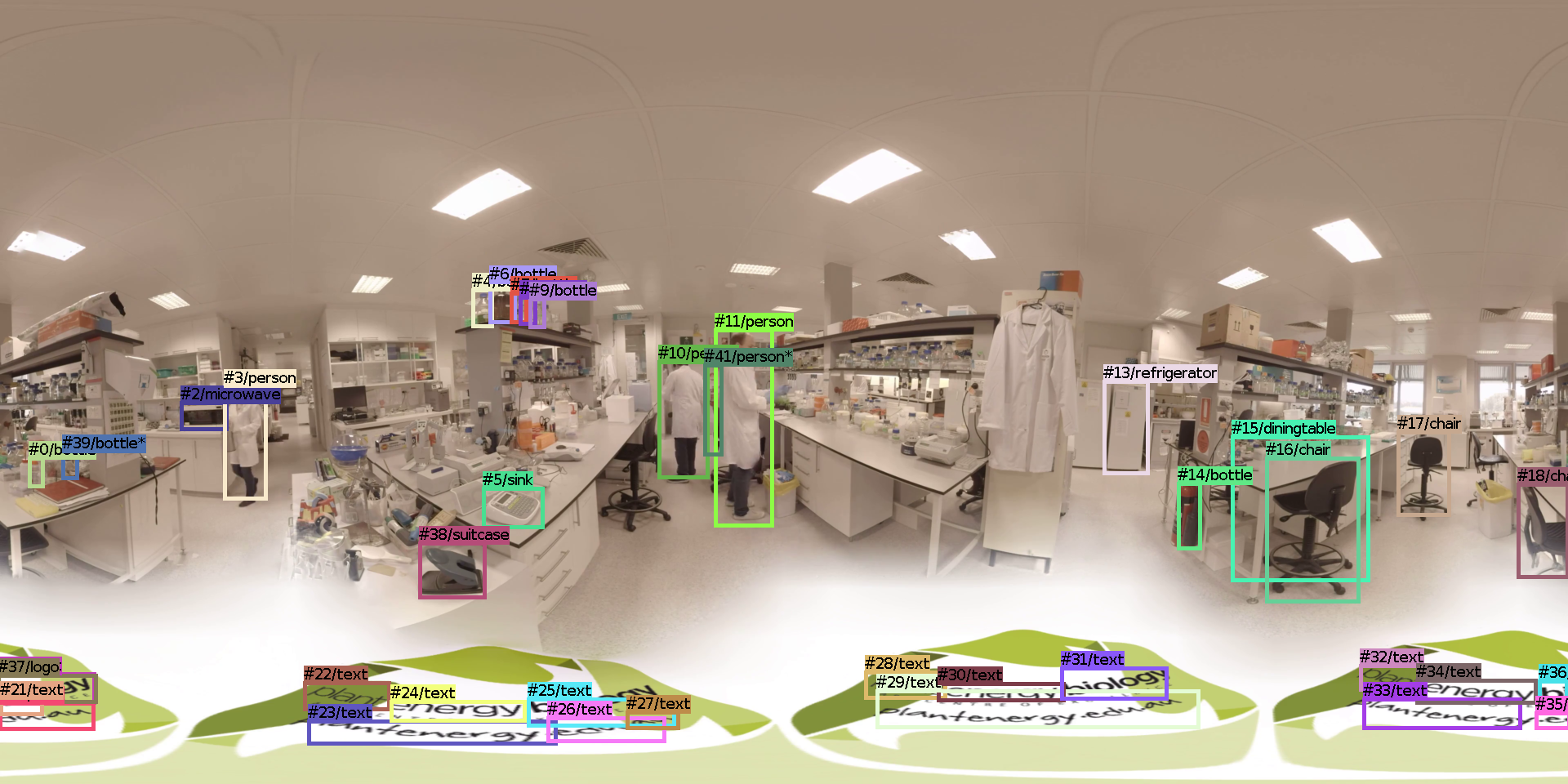}
	\caption{Visualized result of the OmniTrack algorithm for a video of the \emph{Salient360} \cite{Gutirrez2018} dataset.} 
	\label{fig:alg_result_salient360}
\end{figure*}

\begin{figure}[b]
	\centering
		\includegraphics[width=0.4\textwidth]{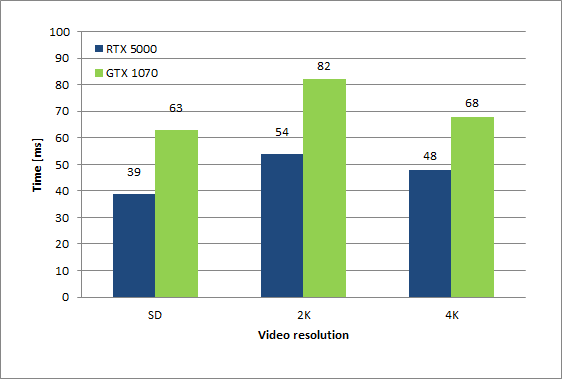}
	\caption{Average runtime of the OmniTrack algorithm for one frame, in different resolutions (SD, 2K, 4K).} 
	\label{fig:runtime_chart}
\end{figure}

For measuring the runtime of the OmniTrack algorithm, we employ a Windows 10 PC with a Xeon E5-1620v3 QuadCore-CPU with 3.50 Ghz. We do the runtime measurement for two different GPUs, a NVIDIA Quadro RTX 5000 and a NVIDIA Geforce 1070 (note that the OmniTrack algorithm uses only one GPU). We measure the average runtime per frame for three different resolutions: Standard Definition (SD, $720x576$), 2K ($2560x11280$) and 4K ($4096x2048$). For SD we employ the original resolution, for 2K pyramid level 1 is used (half resolution) and for 4K pyramid level 2 is used (quarter resolution). In Figure \ref{fig:runtime_chart} the average runtime per frame (in milliseconds) is shown for each resolution. One can see that the OmniTrack algorithm is real-time capable (25 frames per second) for videos in SD resolution on the Quadro RTX 5000 GPU.

Regarding the quality and robustness of the OmniTrack algorithm, we did qualitative tests with a variety of content of different type (static or moving camera, small / large objects, slow / fast motion, film / video,  different impairments like noise / flicker, ...), captured with conventional camera as well as $360^\circ$ cameras. These tests indicate that the algorithm is very robust even for challenging content, likely  due the high quality of the underlying components like
the state of the art object detector and the high-quality optical flow algorithm. In Figure \ref{fig:alg_result_salient360} the result of the OmniTrack algorithm is visualized for a video (in equirectangular projection) of the \emph{Salient360} \cite{Gutirrez2018} dataset, captured with a $360^\circ$ camera.

\section{CONCLUSION}
\label{sec:conclusion}

We presented OmniTrack, an efficient and robust algorithm which is able to automatically detect and track objects, text as well as brand logos. We described the performance optimizations we implemented for the reference YoloV3 object detector and the 
major steps in the training procedure for the combined detector for text and logo. Experiments show that the proposed algorithm runs in real-time for standard definition video. 

\section*{Acknowledgment}

This work has received funding from the European Union's Horizon 2020 research and innovation programme, grant n$^\circ$ 761934, Hyper360 (``Enriching 360 media with 3D storytelling and personalisation elements''). Thanks to Rundfunk Berlin-Brandenburg (RBB) for providing the 360\textbf{$^\circ$}~video content.


\bibliographystyle{abbrv}
\bibliography{conference_paper}

\end{document}